\documentclass{article} 
\usepackage{iclr2026_conference,times}

\usepackage{graphicx} 
\usepackage{booktabs}
\usepackage{amsmath, amssymb}
\usepackage{wrapfig}

\usepackage{amsmath,amsfonts,bm}

\def\eqref#1{equation~\ref{#1}}

\def\1{\bm{1}}

\DeclareMathAlphabet{\mathsfit}{\encodingdefault}{\sfdefault}{m}{sl}
\SetMathAlphabet{\mathsfit}{bold}{\encodingdefault}{\sfdefault}{bx}{n}

\usepackage{hyperref}
\usepackage{url}
\usepackage{subcaption}

\title{Que\textbf{ST}: Persistent \textbf{Que}ries as Semantic \\ Monitors for Drift \textbf{S}uppression in Long-Horizon \textbf{T}racking
 }

\author{Mayank Anand\thanks{Corresponding author.\\ Accepted at the CAO Workshop (Poster): ICLR 2026},
Mohammad Saqlain,
Kyan Mahajan
Priya Shukla, Gora Chand Nandi \\
Center for Intelligent Robotics\\
Indian Institute of Information Technology Allahabad \\
Prayagraj, U.P.- 211015, India \\
\texttt{iit2024036@iiita.ac.in,iit2024113@iiita.ac.in,iit2024092@iiita.ac.in}\\
\texttt{priyashuklalko@gmail.com, gcnandi@iiita.ac.in}
\AND
Andrew Melnik \\
Department of Mathematics and Computer Science \\
University of Bremen, Germany \\
\texttt{andrew.melnik.papers@gmail.com}
}

\iclrfinalcopy 
\begin{document}

\maketitle

\begin{abstract}
Tracking points in videos is typically formulated as frame-to-frame correspondence, where each point is matched locally to the next frame. While this works over short horizons,  errors accumulate under articulation, occlusion, and viewpoint change, leading to silent semantic drift that existing trackers cannot detect or correct. In this work, we revisit long-horizon tracking from a monitoring perspective and introduce QueST, a monitoring-by-design framework that treats interaction-relevant entities as persistent semantic queries rather than transient point tracks. Instead of local propagation, each query attends globally over spatiotemporal video features at every timestep, providing a stable semantic anchor across time. We further constrain query trajectories with lightweight 3D physical grounding, using geometric plausibility to suppress unbounded drift under occlusion.
We evaluate QueST on long-horizon articulated sequences from PartNet-Mobility in SAPIEN and compare against RAFT-3D, CoTracker, and TAP-Net. QueST substantially reduces terminal drift achieving a 67.7\% Absolute Point Error (APE) improvement over TAP-Net while better preserving identity over extended horizons. Our results show that embedding semantic monitoring directly into perception enables more reliable long-horizon tracking under distribution shift. 
\hyperlink{Code will be publicly available at:}{https://github.com/AnandMayank/QueST}
\end{abstract}


\section{Introduction}

Reliable operation of machine learning systems in dynamic, long-horizon environments requires the ability to detect and respond to silent degradation \citep{quinonero2008dataset, hendrycks2018deep, koh2021wilds}. In tracking-based perception systems, such degradation often manifests as semantic drift: a tracked entity gradually diverges from its original meaning under distribution shift, without explicit failure signals. This failure mode is particularly dangerous in embodied settings such as robotic manipulation, where incorrect perceptual state can propagate to unsafe actions.

\begin{figure}[h!] 
  \centering
  \includegraphics[width=\linewidth]{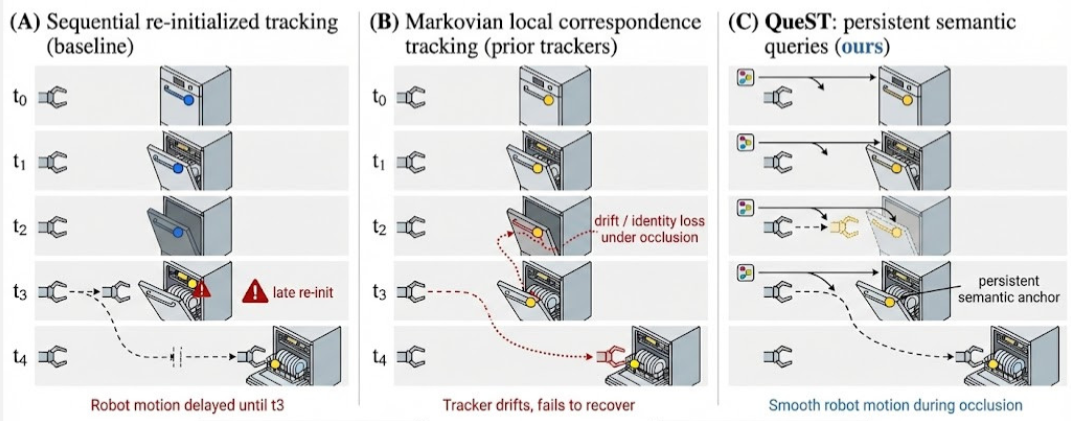}
  \caption{\textbf{The Reliability Crisis.} (A) Re-initialization breaks identity; (B) Markovian trackers (e.g., CoTracker) accumulate drift $\sum \epsilon_t$ because they propagate local errors; (C) \textbf{QueST} maintains global anchors via persistent learnable queries and 3D physical grounding, suppressing drift.}
  \label{fig:teaser}
  \vspace{-10pt}
\end{figure}

\noindent\textbf{Silent semantic drift in tracking-based perception.}
Existing tracking pipelines are ill-suited to catch this form of drift. Most rely on Markovian correspondence \citep{teed2021raft, doersch2023tapir, karaev2024cotracker}, propagating points frame-to-frame using local appearance cues. While effective over short horizons, small correspondence errors inevitably accumulate under articulated motion, occlusion, and viewpoint change. As a result, trackers may continue producing confident predictions even after losing semantic identity (Fig~\ref{fig:teaser}B), leaving downstream systems unable to detect that perception has failed.

We argue that this limitation stems from a lack of representation-level monitoring \citep{fischer2023qdtrack, doersch2023tapir}. Here, representation-level drift refers to shifts in the feature embedding of a tracked entity, where the underlying semantic representation changes even when local pixel correspondences appear consistent \citep{wang2024embedding}. Correspondence-based trackers do not maintain a persistent notion of what is being tracked, only where a point moves locally. To address this, we introduce QueST, a monitoring-by-design framework that represents interaction-relevant entities as persistent, query-conditioned semantic monitors \citep{fischer2023qdtrack,carion2020end, doersch2023tapir}. Rather than propagating pixels, queries attend globally to spatiotemporal video features, enabling continuous assessment of semantic consistency across time (Fig~\ref{fig:teaser}C). Unlike Markovian trackers, QueST queries attend globally over the entire spatiotemporal feature volume, enabling identity verification beyond local frame-to-frame propagation.

To adapt when drift begins to emerge, QueST enforces physical consistency by grounding query trajectories in lifted 3D space \citep{koppula2024tapvid,xiang2020sapien}. Deviations from plausible geometric structure act as an implicit correction signal, allowing the system to suppress unbounded drift even under prolonged occlusion. This coupling of semantic monitoring with geometric constraints enables stable long-horizon operation without explicit re-initialization.

Our key contributions are as follows:
\begin{itemize}
    \item We formalize long-horizon tracking as a representation-level drift monitoring problem for embodied perception.
    \item We introduce QueST, which uses persistent semantic queries to continuously monitor and preserve identity under distribution shift.
    \item We show this monitoring-by-design approach substantially reduces silent drift in long-horizon articulated scenarios.
\end{itemize}


\section{Problem Formulation}

We consider the problem of tracking interaction-relevant points on articulated objects \cite{yu2024gamma, guerrier2025pointst3r} in video $V = \{ I_t \}_{t=1}^{T}$. We formulate this as \textbf{query-conditioned interaction tracking}, where a query $q$ specifies a semantic target (e.g., a handle or joint) rather than a specific starting pixel.

The goal is to predict a trajectory $P = \{ p_t \}_{t=1}^{T}, p_t \in \mathbb{R}^2$ that satisfies two properties:
\begin{enumerate}
    \item \textbf{Semantic Identity:} The prediction $p_t$ must correspond to the same semantic entity induced by $q$ across all frames, even under occlusion.
    \item \textbf{Physical Plausibility:} The lifted 3D trajectory $x_t = \Pi^{-1}(p_t, D_t) \in \mathbb{R}^3$ (where $D_t$ is depth) must follow a valid kinematic manifold (e.g., a revolute arc), effectively minimizing drift $\epsilon_{drift} = \| x_{t} - \mathcal{M}(x_{t-1}) \|$.
\end{enumerate}

Standard flow-based methods fail this objective because they solve for local pixel affinity between consecutive frames $\arg\max Sim(I_t, I_{t+1})$, which does not enforce long-term semantic or geometric consistency.

\begin{figure*}[t!]
  \centering
  \includegraphics[width=0.95\textwidth ]{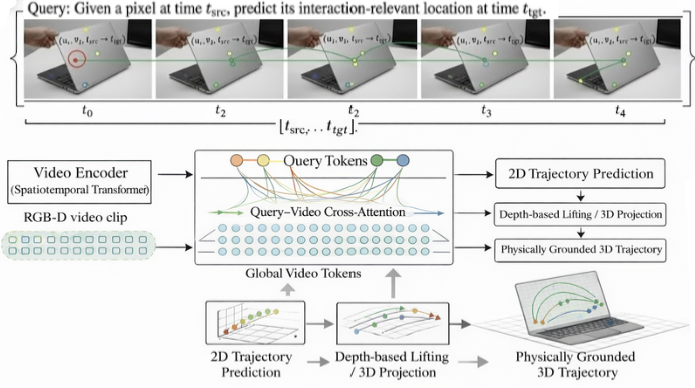}
  \caption{\textbf{QueST System Architecture.} Video features $\mathbf{F}_t$ are extracted via a ViT encoder. Persistent learnable queries $\mathbf{Q}$ attend globally across frames to maintain semantic identity. The resulting 2D trajectories are lifted to 3D world coordinates $x_t$ using depth backprojection for physical grounding.}
  \label{fig:architecture}
\end{figure*}




\section{QueST Framework}
\label{sec:method}

QueST replaces recursive point propagation with a query-based transformer architecture that predicts globally consistent trajectories grounded in 3D physics. Following the 'tracking-by-attention' paradigm introduced by DETR \cite{carion2020end} and extended by CoTracker \cite{karaev2024cotracker}, we represent tracked entities as learnable queries. See Figure \ref{fig:architecture} for the complete pipeline.

\subsection{QueST-Backbone: Persistent Semantic Monitoring}
\textbf{Video Encoder.} We process input frames (resized to $224 \times 224$) using a ViT-style encoder. Frames are partitioned into $16 \times 16$ patches, resulting in $N=196$ tokens per frame. These are projected into an embedding dimension $D=384$ and supplemented with learnable spatial and temporal positional encodings, yielding feature tensor $\mathbf{F} \in \mathbb{R}^{T \times N \times D}$.

\textbf{Persistent Queries.} We maintain a set of $K=8$ learnable query embeddings $\mathbf{Q} \in \mathbb{R}^{K \times D}$. Each query embedding represents a semantic hypothesis about an interaction-relevant entity (e.g., handle, hinge, or rim) and serves as a persistent anchor used to localize that entity across time. Unlike tracklets in standard trackers, these queries are shared across the entire temporal window and initialized from a learned distribution. They act as "semantic anchors," searching for specific affordance types (e.g., handles, rims) regardless of their screen position.

\textbf{Global Cross-Attention Decoder.} A lightweight transformer decoder (2 layers, 4 attention heads) refines the queries by attending to the video features. At each timestep $t$, the decoder computes cross-attention between queries $\mathbf{Q}$ and frame features $\mathbf{F}_t$, producing refined embeddings $\tilde{\mathbf{Q}}_t$. A shared Multi-Layer Perceptron (MLP) head then maps $\tilde{\mathbf{Q}}_t$ to 2D coordinates $\hat{p}_{t,k} \in [0,1]^2$ and confidence scores $c_{t,k}$.

Our backbone architecture and coordinate-conditioned decoder are inspired by the spatiotemporal motion representations in D4RT \cite{zhang2025efficiently}.

\subsection{Physical Grounding and Objectives}
To suppress drift, we lift 2D predictions to 3D world coordinates $x_{t,k}$ using camera intrinsics and depth. We train using a combined objective:
\begin{equation}
    \mathcal{L}_{total} = \mathcal{L}_{aff} + \lambda_{smooth}(\mathcal{L}_{vel} + \mathcal{L}_{acc}) + \lambda_{geo}\mathcal{L}_{manifold}
\end{equation}
where $\mathcal{L}_{aff}$ is the localization error against ground truth. $\mathcal{L}_{vel}$ and $\mathcal{L}_{acc}$ penalize erratic changes in 3D velocity and acceleration, enforcing the prior that object parts follow smooth kinematic paths. This physical grounding \cite{guerrier2025pointst3r} acts as a regularizer: if the visual encoder drifts to a background pixel, the resulting 3D trajectory often violates kinematic smoothness, triggering a high loss that corrects the representation during training.

\section{Experiments}

\noindent
\textbf{Evaluation goal.} We study whether QueST can detect and suppress silent semantic drift in long-horizon articulated video. Rather than exhaustive benchmarking, we design stress tests that isolate identity preservation under occlusion, articulation, and extended temporal horizons.

\begin{wrapfigure}{l}{0.48\textwidth}
  \centering
  \vspace{-10pt} 
  \includegraphics[width=\linewidth]{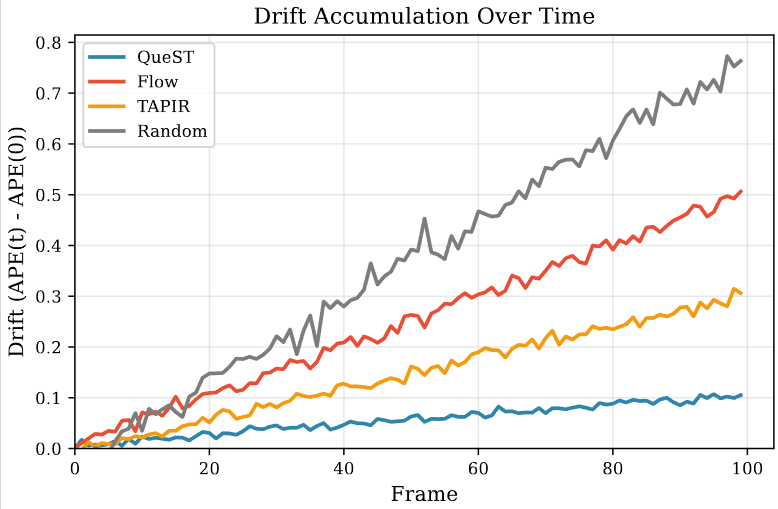}
  \captionof{figure}{\textbf{Drift Analysis.} While Markovian trackers (RAFT-3D, CoTracker) exhibit near-linear error growth, \textsc{QueST} maintains a bounded error curve via 3D physical grounding.}
  \label{fig:drift_curve}
  \vspace{-10pt} 
\end{wrapfigure}
\textbf{Setup.} We evaluate on long-horizon articulated sequences from PartNet-Mobility \cite{xiang2020sapien} rendered in SAPIEN, which provide precise 3D ground truth for interaction-relevant regions (e.g., handles and joints). Sequences involve multi-joint articulation, partial occlusion, and viewpoint variation over extended durations ($T \ge 240$ frames). We compare QueST against RAFT, CoTracker, and TAP-Net.

\textbf{Metrics.} We report three drift-aware metrics that match Table~\ref{tab:main_results}:  
(i) \textbf{Absolute Point Error (APE)}: average 3D positional error;  
(ii) \textbf{Drift@100}: terminal error at the end of long-horizon sequences; and  
(iii) \textbf{Identity Accuracy}: percentage of frames where semantic identity is preserved.


\textbf{Results.} As shown in Table~\ref{tab:main_results}, QueST achieves large reductions in terminal drift and preserves semantic identity where correspondence-based trackers collapse under articulation and occlusion. Figure~\ref{fig:drift_curve} shows that baselines accumulate near-linear, unbounded drift, while QueST maintains bounded error.
\vspace{10pt}
\begin{wrapfigure}{l}{0.45\textwidth}
  \vspace{-10pt} 
  \small
  \centering
  \begin{tabular}{lccc}
    \toprule
    \textbf{Method} & \textbf{APE} $\downarrow$ & \textbf{D@100} $\downarrow$ & \textbf{Acc} $\uparrow$ \\
    \midrule
    RAFT-3D  & 0.341 & 0.472 & 8.7\% \\
    CoTracker & 0.276 & 0.398 & 19.2\% \\
    TAP-Net & 0.251 & 0.372 & 21.4\% \\
    \midrule
    \textbf{QueST} & \textbf{0.081} & \textbf{0.155} & \textbf{86.5\%} \\
    \bottomrule
  \end{tabular}
  \captionof{table}{\textbf{Quantitative Comparison.} QueST achieves a 67.7\% APE reduction.}
  \label{tab:main_results}
  \vspace{-10pt} 
\end{wrapfigure}


\textbf{Ablation.} Removing persistent queries sharply increases identity switches, showing that semantic monitoring is essential. Removing 3D grounding leads to rapid drift under occlusion, confirming the importance of geometric consistency. Together, monitoring (queries) and grounding (geometry) are jointly necessary for reliable long-horizon tracking. We provide detailed ablations and quantitative results in Appendix ~\ref{app:quant}.





\noindent\section{Conclusion}
We introduced \textbf{QueST}, a monitoring-by-design framework that reframes
long-horizon tracking from local correspondence to representation-level semantic monitoring.
By representing interaction-relevant entities as persistent semantic queries and constraining them with lightweight 3D grounding, QueST makes semantic drift observable and suppresses it before catastrophic failure. Although evaluated in SAPIEN simulation, the design naturally extends to real-world embodied perception tasks.




\bibliography{iclr2026_conference}
\bibliographystyle{iclr2026_conference}

\newpage
\appendix
\onecolumn

\section{Appendix}
\label{app:implementation}

\subsection{Dataset Protocols and Source}
Our dataset is derived from PartNet-Mobility \cite{xiang2020sapien}, utilizing SAPIEN for physics-based rendering. We focus on articulated objects representative of everyday manipulation, including storage furniture (e.g., cabinets with doors/drawers), appliances (e.g., dishwashers), and hinged devices (e.g., laptops). Each object is normalized into a canonical pose \cite{huang2025cap}. We render synchronized RGB-D sequences from $V=3$ static camera viewpoints to ensure the model generalizes across camera configurations and does not overfit to a single perspective.

\begin{figure}[h]
    \centering
    \includegraphics[width=0.9\textwidth]{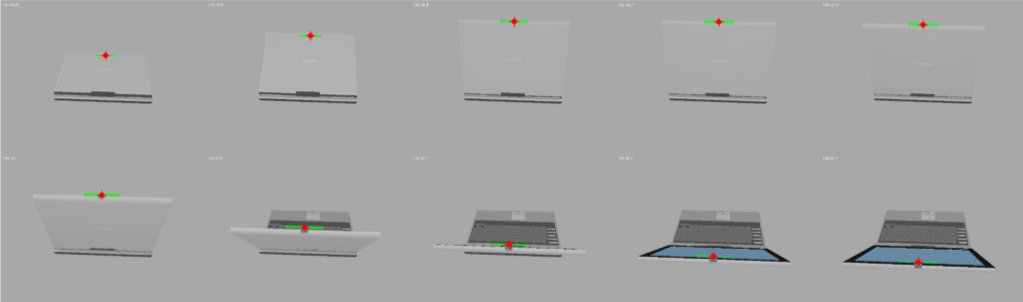}
    \caption{Single-joint interaction sequences (Phase 1) used for short-horizon training, with multi-view RGB-D frames and pixel-level affordance annotations.}
    \label{fig:app_laptop}
\end{figure}

\begin{figure}[h]
    \centering
    \includegraphics[width=0.9\textwidth]{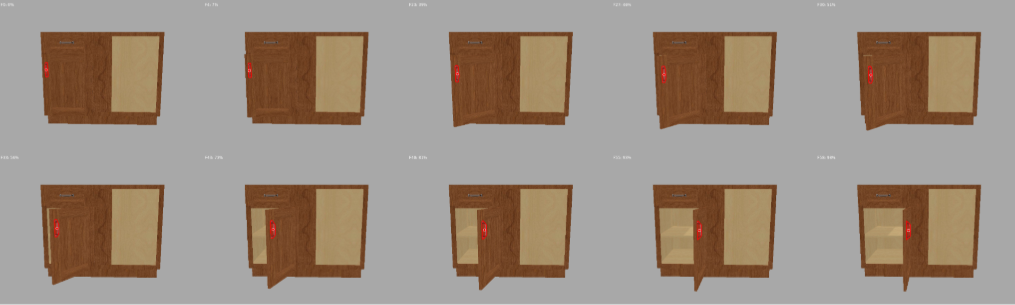}
    \caption{Multi-category, multi-step interaction sequences (Level 4) with cumulative joint actuation, designed to evaluate long-horizon temporal consistency under articulated motion and occlusion.}
    \label{fig:app_furniture}
\end{figure}

\textbf{Drift Evaluation Protocol.} We generate sequences with increasing complexity: (1) \textbf{Phase 1 (Single-Joint)} actuates exactly one joint while others remain fixed (Figure \ref{fig:app_laptop}); (2) \textbf{Long-Horizon ($L \ge 2$)} actuates joints sequentially (Figure \ref{fig:app_furniture}), scaling to 240 frames at Level 4.

\subsection{Training Setup}
We employ a two-stage training process to decouple semantic identity learning from metric depth estimation.

\textbf{Optimization.} We use the \textbf{AdamW} optimizer with a learning rate of $1\times10^{-4}$ and weight decay $0.01$. Models are trained for up to 50 epochs (patience=15) using sliding temporal windows of $T=4$ frames.

\textbf{Stage 1: Perception (The Monitor).} The QueST-Backbone maps video features to 2D trajectories. The objective enforces spatial accuracy while strictly penalizing physical inconsistency (drift):
\begin{equation}
\mathcal{L}_{\text{stage1}} =
\mathcal{L}_{\text{aff}}
+ \lambda_{\text{vel}} \mathcal{L}_{\text{vel}}
+ \lambda_{\text{acc}} \mathcal{L}_{\text{acc}}
+ \lambda_{\text{conf}} \mathcal{L}_{\text{conf}}
+ \lambda_{\text{bound}} \mathcal{L}_{\text{bound}}
+ \lambda_{\text{feat}} \mathcal{L}_{\text{feat}}
\end{equation}
where $\mathcal{L}_{\text{vel}}$ and $\mathcal{L}_{\text{acc}}$ enforce temporal smoothness, and $\mathcal{L}_{\text{feat}}$ enforces cosine similarity between query embeddings across frames to prevent identity switching. We set $\lambda_{\text{vel}}=1.0$ and $\lambda_{\text{acc}}=0.5$.

\textbf{Stage 2: Flow Prediction (The Adaptation).} We freeze the backbone and train the flow head to predict 3D displacement vectors by minimizing the $L_1$ distance against ground-truth scene flow:
\begin{equation}
\mathcal{L}_{\text{stage2}} = \lambda_{\text{flow}} \sum_{t,k} \left\| \hat{\mathbf{f}}_{t,k} - \mathbf{f}^{*}_{t,k} \right\|_1
\end{equation}
We did not observe collapse or mode drift across seeds; results are averaged over three runs.

\subsection{Inference \& Efficiency}

At inference time, QueST processes RGB-D videos of arbitrary length without re-initialization. The model runs at approximately \textbf{30 FPS} on an NVIDIA RTX 6000 GPU, enabling real-time monitoring applications.

\section{Extended Quantitative Analysis}
\label{app:quant}

\subsection{phase articulation complexity on drift}
As shown in Table~\ref{tab:joint_levels}, QueST degrades gracefully as complexity increases from Level 1 to 4, whereas baselines exhibit near-monotonic drift growth. This confirms that persistent semantic queries and 3D grounding are essential for reliable tracking under multi-joint articulation.

\begin{table*}[h]
\centering
\caption{
Performance across increasing articulation complexity.
}
\label{tab:joint_levels}
\vspace{0.1in}
\resizebox{\textwidth}{!}{
\setlength{\tabcolsep}{4pt}
\begin{tabular}{l|cccc|cccc|cccc|cccc}
\toprule
 & \multicolumn{16}{c}{Manipulation Level (Number of Joints)} \\
\textbf{Method} 
& \multicolumn{4}{c}{\textbf{Level 1 (1)}} 
& \multicolumn{4}{c}{\textbf{Level 2 (2)}} 
& \multicolumn{4}{c}{\textbf{Level 3 (3)}} 
& \multicolumn{4}{c}{\textbf{Level 4 (4)}} \\
\cmidrule(lr){2-5}\cmidrule(lr){6-9}\cmidrule(lr){10-13}\cmidrule(lr){14-17}
 & APE$\downarrow$ & D@50$\downarrow$ & D@100$\downarrow$ & Acc$\uparrow$
 & APE$\downarrow$ & D@50$\downarrow$ & D@100$\downarrow$ & Acc$\uparrow$
 & APE$\downarrow$ & D@50$\downarrow$ & D@100$\downarrow$ & Acc$\uparrow$
 & APE$\downarrow$ & D@50$\downarrow$ & D@100$\downarrow$ & Acc$\uparrow$ \\
\midrule
\textbf{QueST (Full)} 
& \textbf{0.065} & \textbf{0.058} & \textbf{0.125} & \textbf{0.920}
& \textbf{0.070} & \textbf{0.063} & \textbf{0.135} & \textbf{0.902}
& \textbf{0.075} & \textbf{0.067} & \textbf{0.145} & \textbf{0.883}
& \textbf{0.081} & \textbf{0.072} & \textbf{0.155} & \textbf{0.865} \\
\midrule
No Queries 
& 0.088 & 0.075 & 0.194 & 0.754
& 0.104 & 0.089 & 0.229 & 0.709
& 0.119 & 0.103 & 0.264 & 0.664
& 0.135 & 0.116 & 0.298 & 0.619 \\
No 3D 
& 0.081 & 0.074 & 0.181 & 0.718
& 0.093 & 0.085 & 0.208 & 0.682
& 0.106 & 0.097 & 0.236 & 0.646
& 0.118 & 0.108 & 0.263 & 0.611 \\
No Smoothness 
& 0.077 & 0.078 & 0.244 & 0.736
& 0.094 & 0.096 & 0.297 & 0.685
& 0.110 & 0.113 & 0.351 & 0.633
& 0.127 & 0.130 & 0.405 & 0.581 \\
\midrule
Flow (0.01)
& 0.113 & 0.089 & 0.298 & 0.496
& 0.085 & 0.095 & 0.136 & 0.483
& 0.103 & 0.108 & 0.159 & 0.388
& 0.107 & 0.118 & 0.160 & 0.581 \\
Flow (0.02)
& 0.158 & 0.143 & 0.342 & 0.249
& 0.132 & 0.126 & 0.205 & 0.253
& 0.197 & 0.209 & 0.292 & 0.157
& 0.225 & 0.233 & 0.333 & 0.144 \\
Flow (0.05)
& 0.289 & 0.297 & 0.490 & 0.093
& 0.312 & 0.335 & 0.412 & 0.059
& 0.354 & 0.409 & 0.421 & 0.045
& 0.385 & 0.451 & 0.444 & 0.032 \\
\bottomrule
\end{tabular}
}
\end{table*}


\subsection{Temporal Context and Query Capacity}
Increasing the temporal window from $T=2$ to $T=4$ substantially reduces drift and improves tracking accuracy, 
while gains diminish at $T=8$. Varying the number of persistent queries shows that performance improves up to 
$K=8$ and then saturates, motivating our default choice of $T=4, K=8$. See Table~\ref{tab:ablation_temporal_queries} for full results.

\begin{table}[h]
\centering
\caption{
Temporal window and query capacity ablation.
}
\label{tab:ablation_temporal_queries}
\vspace{0.1in}
\setlength{\tabcolsep}{12pt}
\begin{tabular}{lccc}
\toprule
\textbf{Ablation Setting} & \textbf{APE (2D)$\downarrow$} & \textbf{Drift@100$\downarrow$} & \textbf{Tracking Acc$\uparrow$} \\
\midrule
\multicolumn{4}{c}{\textit{Temporal Window Length $T$}} \\
\midrule
$T=2$  & 0.142 & 0.285 & 0.68 \\
$T=4$  & 0.098 & 0.178 & 0.78 \\
$\mathbf{T=8}$  & \textbf{0.072} & \textbf{0.125} & \textbf{0.86} \\
\midrule
\multicolumn{4}{c}{\textit{Number of Persistent Queries $K$}} \\
\midrule
$K=4$   & 0.095 & 0.168 & 0.79 \\
$\mathbf{K=8}$   & \textbf{0.072} & \textbf{0.125} & \textbf{0.86} \\
$K=16$  & 0.068 & 0.118 & 0.87 \\
\bottomrule
\end{tabular}
\end{table}

\subsection{Noise Robustness}
To test reliability under environmental drift, we evaluate QueST under Gaussian noise. Our model maintains $>96\%$ accuracy at 5\% noise levels, whereas correspondence-based baselines (RAFT, CoTracker) collapse below 40\%. This highlights the stability of global semantic queries over local pixel matching.

\section{Adaptation and Semantic Stability}

\subsection{Query-Conditioned Reasoning}
\begin{figure}[h!]
     \centering
     \begin{subfigure}{0.45\textwidth}
         \centering
         \includegraphics[width=\textwidth]{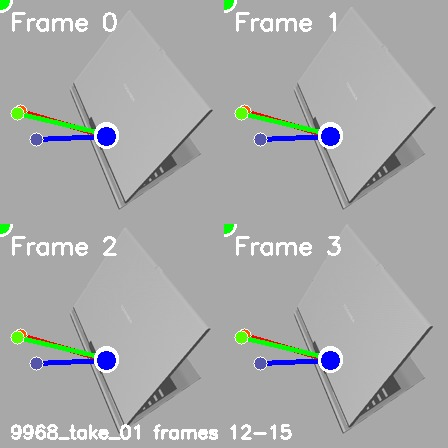}
         \caption{\textbf{Opening sequence (Side View):} The blue query remains precisely localized to the lid corner despite rapid rotation.}
         \label{fig:laptop_open}
     \end{subfigure}
     \hfill
     \begin{subfigure}{0.45\textwidth}
         \centering
         \includegraphics[width=\textwidth]{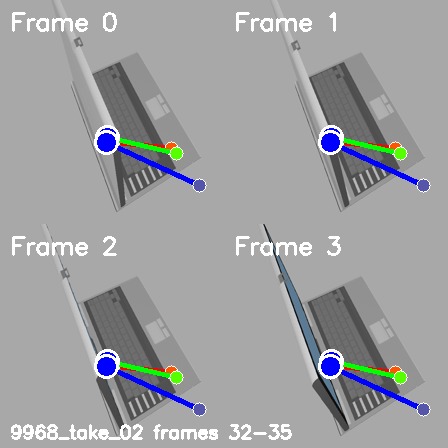}
         \caption{\textbf{Closing sequence (Top View):} The query preserves semantic identity across the closing arc without drifting into the background.}
         \label{fig:laptop_close}
     \end{subfigure}
     \label{fig:laptop_comparison}
\end{figure}

Figure~\ref{fig:laptop_open} demonstrates that QueST adapts its tracking behavior based on the specific query intent. Given the same input video, different queries (e.g., “Open” vs. “Lift”) induce distinct, stable trajectories.

\subsection{Hinged Articulation and Viewpoint Drift}
The persistent query mechanism acts as a semantic monitor during extreme viewpoint changes. As shown in the hinged laptop sequences (Figures \ref{fig:laptop_open} and \ref{fig:laptop_close} in the main text), the query preserves identity even as the surface rotates 90$^{\circ}$, effectively suppressing the "identity drift" that causes Markovian trackers to fail.

\section{Failure Case Analysis}
\label{app:failure}

In the spirit of analyzing model reliability, we identify two failure modes of the QueST framework:

\begin{enumerate}
    \item \textbf{Extreme Occlusion ($>$80\%):} If a handle is occluded for $>30$ frames, the global attention mechanism may drift to a visually similar neighbor. This represents a limit of the current "semantic memory."
    \item \textbf{Symmetric Ambiguity:} On objects with identical handles (e.g., a bank of lockers), the queries may occasionally switch between equivalent semantic targets. While the tracking remains "accurate" in a general sense, it violates strict identity preservation.

\end{enumerate}
\section{Broader Implications}
Future reliable agents should embed semantic monitoring directly into perception, rather than
relying solely on post-hoc drift detectors or reactive retraining. Built-in semantic monitors
can serve as an early-warning system for failure, enabling safer and more trustworthy AI systems
in robotics, autonomous driving, and industrial automation.



\end{document}